\documentclass[10pt, a4paper]{article}

\usepackage{graphicx}
\usepackage{hyperref}
\usepackage{color}
\usepackage[table]{xcolor}
\usepackage{multirow}
\usepackage{booktabs}

\usepackage[english]{babel}
\usepackage{amssymb,amsmath,amsthm, amsfonts}

\RequirePackage[left=2cm,%
                right=2cm,%
                top=2.25cm,%
                bottom=2.25cm,%
                headheight=12pt,%
                letterpaper]{geometry}%

\theoremstyle{definition}

\providecommand{\keywords}[1]{\textbf{\textit{keywords ---}} #1}

\title{Discrete Schroedinger Transform For Texture Recognition}

\author{Jo\~ao B. Florindo$^{1,2}$, Odemir M. Bruno$^{1}$}
\date{}
\begin{document}
\maketitle
\begin{center}
\noindent{$^1$S\~{a}o Carlos Institute of Physics, University of S\~{a}o Paulo, PO Box 369, 13560-970, S\~{a}o Carlos, SP, Brazil.\\Scientific Computing Group - http://scg.ifsc.usp.br}\\
\noindent{$^2$Institute of Mathematics, Statistics and Scientific Computing - University of Campinas\\
	Rua S\'{e}rgio Buarque de Holanda, 651, Cidade Universit\'{a}ria "Zeferino Vaz" - Distr. Bar\~{a}o Geraldo, CEP 13083-859, Campinas, SP, Brasil}
\end{center}

\begin{abstract}
\noindent{This work presents a new procedure to extract features of grey-level texture images based on the discrete Schroedinger transform. This is a non-linear transform where the image is mapped as the initial probability distribution of a wave function and such distribution evolves in time following the Schroedinger equation from Quantum Mechanics. The features are provided by statistical moments of the distribution measured at different times. The proposed method is applied to the classification of three databases of textures used for benchmark and compared to other well-known texture descriptors in the literature, such as textons, local binary patterns, multifractals, among others. All of them are outperformed by the proposed method in terms of percentage of images correctly classified. The proposal is also applied to the identification of plant species using scanned images of leaves and again it outperforms other texture methods. A test with images affected by Gaussian and ``salt \& pepper'' noise is also carried out, also with the best performance achieved by the Schroedinger descriptors.} 
\end{abstract}
\begin{center}
\end{center}

\section{Introduction}

Texture analysis has been a fundamental research topic, being used for image retrieval, object recognition, image segmentation, and so on \cite{MS98}. The developed methods have been successfully used in a number of different fields of application, such as Medicine \cite{RCSHF14}, Biology \cite{MNCREBDGPCR15}, Engineering \cite{HBRMG14}, etc.

Several approaches have been described in the literature to extract meaningful information from texture images \cite{MS98} and a particular category of these methods comprise those employing some sort of image transform, like Fourier \cite{GW02}, wavelets \cite{RBCDMMR95}, and Gabor \cite{ZFB14}, for example. Most of these transforms were primarily developed for image processing, and hence they have the property of describing the image from another viewpoint, clarifying patterns that were not evident by the simple inspection of pixel values.

While transforms like Fourier and wavelets are classical and well-known in the literature, in the last years other paradigms of image transform have been proposed and studied, mainly for applications in image processing. One of such paradigms is the discrete Schroedinger transform, as described in \cite{LZXLG12}. This is a non-linear operation that simulates a quantum system, where the image is assumed to be the initial distribution of a wave function and from that point it starts to evolve following the well-established Schroedinger equation. Despite the promising results achieved by this transform in different tasks of image processing, up to our knowledge, there is no attempt to employ it as an auxiliary operation for the extraction of texture descriptors. 

Based on this, we propose to employ the discrete Schroedinger transform for the extraction of meaningful features capable of expressing the information enclosed within a texture in a meaningful way, with precision and robustness. After being operated by the Schroedinger transform, the central statistical moments, as described in \cite{P84}, are extracted from the distribution evolving in time. As detailed for example in \cite{P84}, central moments succinctly quantify statistical distributions, yielding to a powerful description of the transform. 

In this context, the Schroedinger transform works like a non-linear operator over the image, capable of detecting discontinuities in the distribution of pixels, both with regards to the magnitude as well as to the spatial distribution of such discontinuities at different scales. Such heterogeneous patterns play an important role in the localization of structural patterns along the texture and when the statistical information of this operation is provided by the moments, instead of only accounting for the homogeneity of pixels, now the distribution of discontinuities is also quantified, attenuating the redundant information of regions with homogeneous pixels.

The efficiency of the proposed method is assessed on the classification of three texture databases in the literature, to know, Outex \cite{OjalaMPVKH02}, UIUC \cite{KA11} and USPTex \cite{BackesB10}. The results are compared to other well-known texture descriptors and confirm the potential of the proposal to characterise real-world textures. These tests are also repeated when UIUC images are affected by two types of random noise (Gaussian and ``salt \& pepper''). Schroedinger features obtained the best performance in this task once more. Finally, our proposal is applied to a problem from the real world, to know, the identification of plant species based on leaf images. Again, the proposed features achieved the highest rate of samples correctly identified.

\section{Schroedinger Transform}

\subsection{Historical Context}

There are essentially two operations in imaging that led to the development of Schroedinger Transform: distance transforms and image segmentation.

The development of a Schroedinger Distance Transform (SDT) comes from the relation between the Schroedinger equation and Hamilton-Jacobi solvers, used as a generalied model for distance mappings \cite{S12}. This application is proposed in \cite{TMLL09}, starting from the fact that Classical Mechanics is actually only a limiting case of Quantum Mechanics and using Hamilton-Jacobi formalism to find an efficient approximate solution to the Euclidean Distance Transform. Following a similar idea, a wave function representation of a distance transform is developed in \cite{BKLS11} and employed for shape representation. In \cite{S12}, the authors extend the Schroedinger Distance Transform to be used in convolution operations as well as to obtain histograms of oriented gradients (HOG).

Parallel to SDT, another approach named Discrete Schroedinger Transform (DST), applied to image segmentation, is introduced in \cite{LZFD08}. In that work, the authors propose a new method for boundary extraction by adapting a deformable model from Classical Mechanics to a Quantum Mechanics context. This approach is detailed in \cite{LZXLG12} where the transform is generalised as a tool for image analysis, not only for contour extraction, but also as a band-pass filter and for image segmentation. The Schroedinger equation is also used in a similar way in \cite{AKG13} for segmentation, by replacing the potential function by the grey-level image.

Finally, concerning to image processing, it is worth to mention the existence of other quantum-based approaches, like that developed in \cite{E02}, where a Quantum Signal Processing (QSP) framework is proposed, making use of the probabilistic nature of Quantum Mechanics to develop generalised algorithms for signal and image analysis. Although QSP shares some background on Quantum Physics with Schroedinger transform, they are quite different in terms of algorithms and operations.

Here, we focus on the Discrete Schroedinger Transform, using the deformable model approach to Schroedinger transform, as it was already applied to grey-level images (for filtering) and has an easier and more evident generalization to texture analysis.

\subsection{Definition}

The development of the Discrete Schroedinger Transform (DST) in \cite{LZFD08} starts from the analogy between the contour of an object of interest and the path of a particle. A well-known model \cite{TWK88} in image segmentation states that the parametrised contour $X(s) = (x(s),y(s)) : s \in [0,1]$ of a real-world object should minimise the following energy functional:
\begin{equation}\label{eq:energy}
	E = \int_{0}^{1}{\frac{1}{2}(w_1|X'(s)|^2 + w_2|X''(s)|^2) + E_o(X(s)) ds},
\end{equation}
where $w_1$, $w_2$ and $E_o$ are parameters named, respectively, tension, rigidity and outer energy. All of them are tuned according to each particular problem. 

The function minimizing $E$ can be seen actually as the shortest path between $s=0$ and $s=1$, satisfying the underlying constraints. Quantum Mechanics generalises such idea by replacing $X(s)$ by a particle moving from the initial position $a$ at a time $t_a$ towards the final position $b$ at time $t_b$. Now the concept of minimum distance is replaced by what is called a kernel $K(b,a)$ and this takes into consideration not only the shortest but all the possible paths between $a$ and $b$, appropriately weighted. $K(b,a)$ is computed by summing up the contribution of each path $X(t)$, according to
\begin{equation}
	K(b,a) = \sum_{X \in P_{ab}}{\phi(X(t))},
\end{equation}
where $P_{ab}$ is the set of all possible paths between $x_a$ and $x_b$ and $\phi(X(t))$ is given by
\begin{equation}
	\phi(X(t)) = Ce^{(j 2 \pi /h)E(X(t))},
\end{equation}
being $C$ and $h$ constants, $j$ the imaginary number, and $E$ obtained by minimizing the generalised energy functional, given by
\begin{equation}
	E = \int_{t_a}^{t_b}{L(X'(t),X(t),t)dt},
\end{equation}
where $L$ is the Lagrangian of the system. The probability of a particle moving from $a$ to $b$ is given by:
\begin{equation}
	P(b,a) = |K(b,a)|^2.
\end{equation}

$P(b,a)$ (and $K(b,a)$ as a consequence) need to be computed in order to find the law of motion for the particle and therefore the minimizing function, which by recalling the analogy to the energy function (Equation \ref{eq:energy}), will be useful for image analysis purposes. Although this is a solvable problem when the Lagrangian is well-known and simple, it becomes very difficult and time-consuming for more complex systems. In such cases, numerical approximations use to be computed, starting from the representation of the kernel by a wave function $w(x,t)$ solving the Schroedinger equation:
\begin{equation}
	j\overline{h}\frac{\partial w}{\partial t} = -\frac{\overline{h}^2}{2m} \left( \frac{\partial^2 w}{\partial x^2} + \frac{\partial^2 w}{\partial y^2} \right) +V(x,t)w(x,t),
\end{equation}
where $\overline{h}$ and $m$ are constants and $V(x,t)$ is the potential component in the Lagrangian of the system. In the domain of an image $I(x,y)$ two types of representation for this equation can be defined, although here we focus on the simplest I-type Schroedinger transform, where the potential part is discarded, and the equation is represented in a compact version through:
\begin{equation}\label{eq:schroedinger_image}
	\left\{
		\begin{array}{c}
			j w_t + k \triangle w = 0\\
			w \mid _{t=0} = I(x,y)
		\end{array}
	\right.,
\end{equation}
where $\triangle$ is the Laplacian operator and $k$ is a constant.

The numerical solution $w$ of Equation \ref{eq:schroedinger_image} can be found in a simple and fast way through the Fourier transform. In the Fourier domain, the transform $\hat{w}(I)$ is given by
\begin{equation}\label{eq:Schroed_fourier}
	\hat{w}(I) = e^{-jkd_{uv}t}\hat{I}(u,v),
\end{equation}
where $\hat{I}(u,v)$ is the Fourier transform of $I(x,y)$ and $d_{uv}$ is a mapping of Euclidean distances from the centre of the frequency domain in an $m \times n$ mask:
\begin{equation}
	d_{uv} = (u-m/2)^2 + (v-n/2)^2.
\end{equation}

Finally, the Schroedinger Transform  $S(I)$ of the image is given, as can be supposed, by the modulus of the Fourier inverse transform of $\hat{w}(I)$. The parameter $k$ is free to be set according to each particular application.

\section{Proposed Method}

This work proposes the use of the discrete Schroedinger transform to obtain image descriptors, with particular application to the classification of grey-scale texture images. DST has been applied to several problems in image processing and analysis, like filtering, segmentation, contour extraction, etc. \cite{LZXLG12,S12}. However, up to our knowledge, it has never been applied before to extract image features.

\begin{figure}[!htpb]
	\centering	
	\begin{tabular}{cc}
		\includegraphics[width=.5\textwidth]{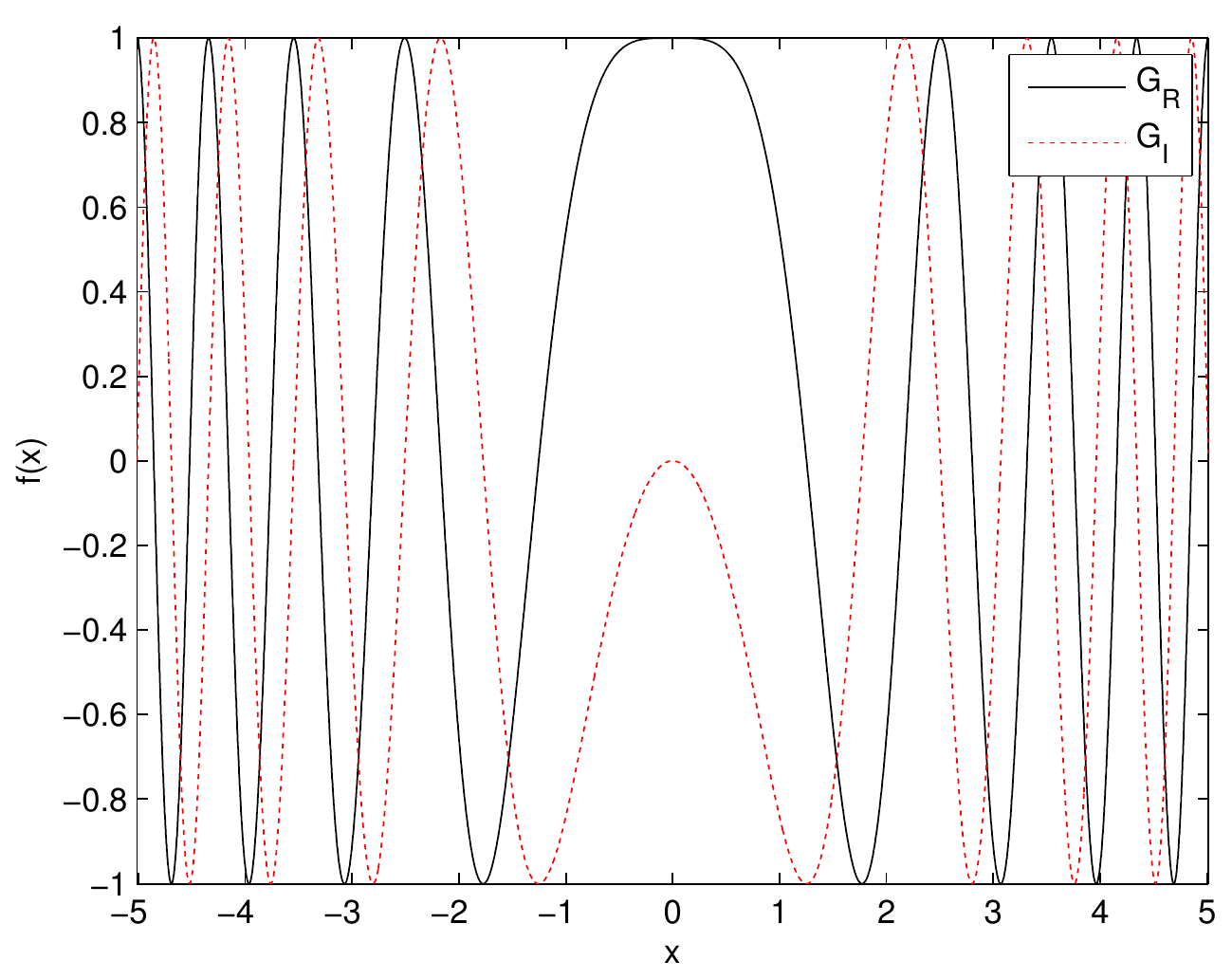} &
		\includegraphics[width=.5\textwidth]{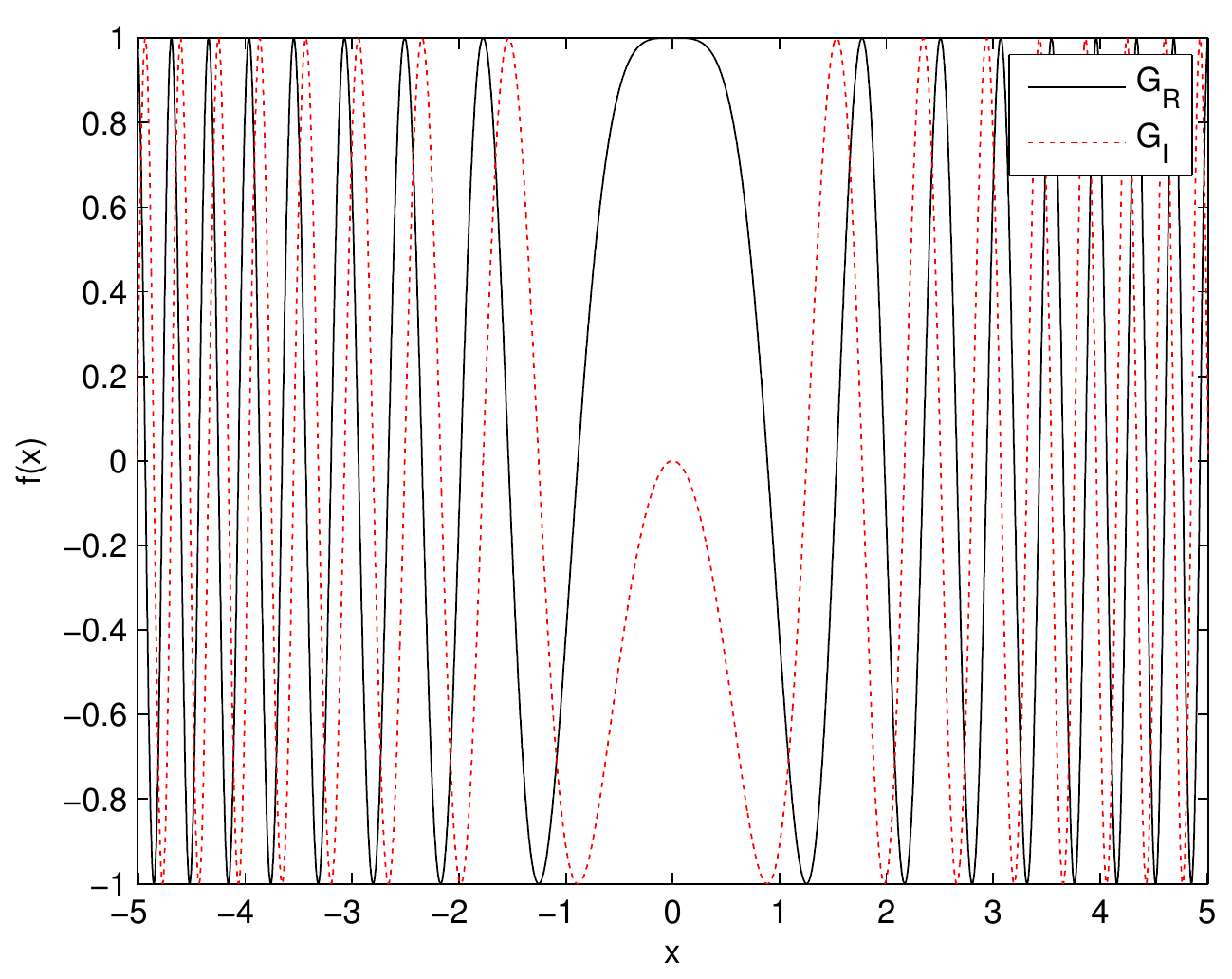} \\
		$t=1$ & $t=2$\\
	\end{tabular}				
	\caption{Real and imaginary part of the kernel $G$ for $t=1$ and $t=2$.}
	\label{fig:mask}                                  
\end{figure}

More specifically, we analyse the time evolution of the Schroedinger equation when the initial wave function is assumed to correspond to the original image. We know from the theory of Fourier transforms that Equation \ref{eq:Schroed_fourier} is equivalent to a convolution in spatial domain with the Fourier inverse transform of a complex Gaussian, which is also known to be another complex Gaussian. As we are not concerned with absolute values, any constant can be disregarded and therefore we end up working with the simplified kernel $G$:
\begin{equation}
	G = e^{-jtx^2},
\end{equation}
where $x$ is the spatial coordinate and $t$ is now a basic parameter (originally related to time in the quantum physics context). This expression can be trivially broken into its real ($G_R$) and imaginary ($G_I$) part:
\begin{equation}\label{eq:def1}
	G_R = \cos(tx^2); G_I = -\sin(tx^2).
\end{equation} 
These are sinusoids whose frequencies increase quadratically and Figure \ref{fig:mask} depicts $G_R$ and $G_I$ for $x \in [-100,100]$ and for two different values of $t$.

To facilitate some basic computations, we assume that the function to be transformed is a one-dimensional profile (potentially the cross-section of a real image), represented by a function $f(x)$, where $x$ is the underlying coordinate. The convolution of $f \ast G$ is given by
\begin{equation}	
	f \ast G = f \ast G_R + j(f \ast G_I).
\end{equation}
We are especially interested on the magnitude of such convolution. To avoid some unnecessarily cumbersome calculus involving square roots and based on the monotonicity of the quadratic function, we will focus on the squared magnitude:
\begin{equation}\label{eq:convSq}
	|f \ast G|^2 = (f \ast G_R)^2 + (f \ast G_I)^2.
\end{equation}
Now we recall the expression for the discrete convolution of $f(x)$ with a generic finite-supported function $g(x)$ whose support domain lies within $[-r,+r]$:
\begin{equation}\label{eq:def2}
	(f \ast g)(x) = \sum_{i=-r}^{+r}{f(x+i)g(i)}
\end{equation} 
Plugging \ref{eq:def1} and \ref{eq:def2} into \ref{eq:convSq} we have
\begin{equation}
	|f \ast G|^2(x) = 	\left( \sum_{i=-r}^{+r}{f(x+i)\cos(ti^2)} \right) ^2 + \left( \sum_{i=-r}^{+r}{f(x+i)\sin(ti^2)} \right)^2
\end{equation}
Using the parity of $i^2$ for positive and negative $i$ and the well-known values $\cos(0)=1$ and $\sin(0)=0$, a further development can be achieved:
\begin{equation}
	\begin{array}{ll}
	|f \ast G|^2(x) = & \left( f(x) + \sum_{i=1}^{r}(f(x-i)+f(x+i))\cos(ti^2) \right)^2 + \\
	& \left(\sum_{i=1}^{r}(f(x-i)+f(x+i))\sin(ti^2) \right)^2, 
	\end{array}
\end{equation}
which, after the definition of $d_i = f(x-i)+f(x+i)$, can be written in a shorter equation:
\begin{equation}\label{eq:conv_short}
|f \ast G|^2(x) = \left( f(x) + \sum_{i=1}^{r}d_i\cos(ti^2) \right)^2 + \left(\sum_{i=1}^{r}d_i\sin(ti^2) \right)^2, 
\end{equation}
Given that the above expression involves trigonometric functions and a squared summation, its development is clarified by using trigonometric relations and the multinomial theorem, according to which
\begin{equation}
	(x_1 + x_2 + ... + x_m)^2 = \sum_{k_1+k_2+...+k_m = 2}\frac{2!}{k_1!k_2!...k_m!}\prod_{1 \leq t \leq m}x_t^{k_t},
\end{equation}
where $k_1, k_2, ..., k_m$ are non-negative integers, that is, for squared expressions they can only be $0$, $1$ or $2$. In practice, this means that
\begin{equation}\label{eq:multinomial}
	(x_1 + x_2 + ... + x_m)^2 = \sum_{i=1}^{m}x_i^2 + 2\sum_{i=1}^{m}\sum_{j=i+1}^{m}x_ix_j.
\end{equation}
Now, Equation \ref{eq:conv_short} is divided into two parts after developing the first square and rearranging:
\begin{equation}
	\begin{array}{ll}
	|f \ast G|^2(x) = & \left[ f^2(x) + 2f(x)\sum_{i=1}^{r}d_i\cos(ti^2) \right] + \\ & \left[ \left( \sum_{i=1}^{r}d_i\cos(ti^2) \right)^2 + \left(\sum_{i=1}^{r}d_i\sin(ti^2) \right)^2 \right]
	\end{array}
\end{equation}
and (\ref{eq:multinomial}) is applied to the rightmost part:
\begin{equation} \label{eq:kernel_developed}
\begin{array}{ll}
|f \ast G|^2(x) = & 
\left[ f^2(x) + 2f(x)\sum_{i=1}^{r}d_i\cos(ti^2) \right] + \\
& \left[ \sum_{i=1}^{r}d_i^2\cos^2(ti^2) + 2\sum_{i=1}^{r}\sum_{j=i+1}^{r}d_id_j\cos(ti^2)\cos(tj^2) + \right.\\ 
& \left. \sum_{i=1}^{r}d_i^2\sin^2(ti^2) + 2\sum_{i=1}^{r}\sum_{j=i+1}^{r}d_id_j\sin(ti^2)\sin(tj^2)  \right]
\end{array}
\end{equation}
To check the validity of the above expression we set $t=0$ and obtain:
\begin{equation}
	|f \ast G|^2(x) = 
	f^2(x) + 2f(x)\sum_{i=1}^{r}d_i + \sum_{i=1}^{r}d_i^2 + 2\sum_{i=1}^{r}\sum_{j=i+1}^{r}d_id_j,
\end{equation}
which we know from the square of the sum formula that is equivalent to
\begin{equation}
	|f \ast G|^2(x) = 
	f^2(x) + 2f(x)\sum_{i=1}^{r}d_i + \left( \sum_{i=1}^{r}d_i \right)^2,
\end{equation}
which in turn is the same as
\begin{equation}
	|f \ast G|^2(x) = 
	\left( f(x)+\sum_{i=1}^{r}d_i \right)^2,
\end{equation}
or, recalling the definition of $d_i$:
\begin{equation}
	|f \ast G|(x) = \sum_{i=-r}^{r}f(x+i),
\end{equation}
making the kernel to act like a neuter element in \ref{eq:def2} and preserving the basic structure of the original function as expected from the initial condition of a physical system.

By grouping terms in (\ref{eq:kernel_developed}) and using the relations $\cos^2(x)+\sin^2(x)=1$ and $\cos(a)\cos(b)+\sin(a)\sin(b) = \cos(a-b)$, we end up with
\begin{equation}
	|f \ast G|^2(x) = 
	f^2(x) + 2f(x)\sum_{i=1}^{r}d_i\cos(ti^2) + \\
	\sum_{i=1}^{r}d_i^2 + 2\sum_{i=1}^{r}\sum_{j=i+1}^{r}d_id_j\cos(ti^2-tj^2).
\end{equation}
By putting $d_i$ in evidence and knowing that $\cos \left( \frac{\pi}{3} \right) =\frac{1}{2}$:
\begin{equation} \label{eq:conv_short2}
	|f \ast G|^2(x) = f^2(x) + 2\sum_{i=1}^{r} d_i \left( \cos(ti^2)f(x) + \cos \left( \frac{\pi}{3} \right) d_i + \sum_{j=i+1}^{r} \cos(ti^2 - tj^2)d_j \right) 
\end{equation}

Observing the above expression it can be noticed that the transform depends in a somewhat complex manner on the value of the function at $x$ and on the neighbourhood expressed by $d_i$ and $d_j$. To analyse the impact of the neighbourhood in the weighted summation, we preserve only terms containing $d_i^2$ and $d_id_j$ in the rightmost part of (\ref{eq:conv_short2}), therefore yielding the expression $\mathfrak{S}_i(j)$:
\begin{equation}
	\mathfrak{S}_i(j) = \sum_{i=1}^{r} d_i \left( \cos \left( \frac{\pi}{3} \right) d_i + \sum_{j=i+1}^{r} \cos(ti^2 - tj^2)d_j \right).
\end{equation}
For $t \ll 1$ the weights provided by $\cos(ti^2 - tj^2)$ present quadratic periodic behaviour as that shown in Figure \ref{fig:mask}. The pattern of the weight becomes especially complex and non-linear provided that only integer values are used for $i$ and $j$.

\begin{figure}[!htpb]
	\centering
	\begin{tabular}{|c|}
    	\hline
		\includegraphics[width=\textwidth]{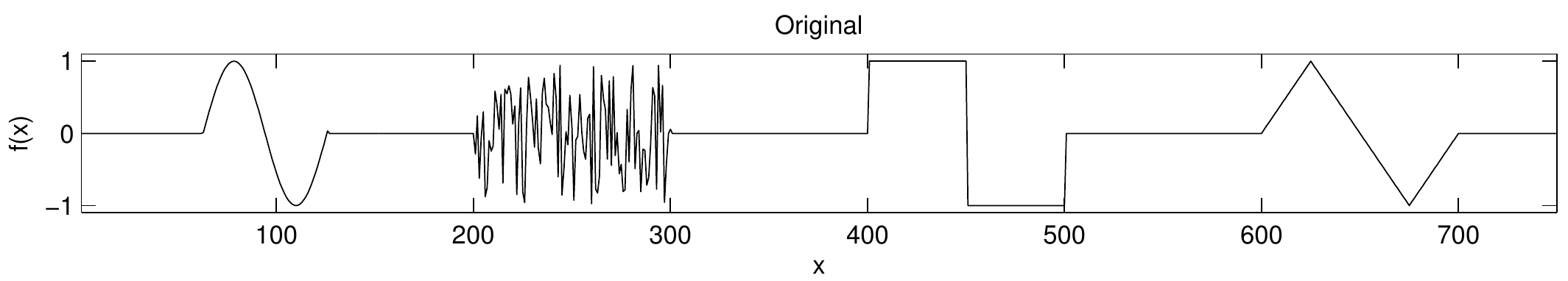}\\
        \hline
    \end{tabular}
    \begin{tabular}{|c|}
    	\hline
		\includegraphics[width=\textwidth]{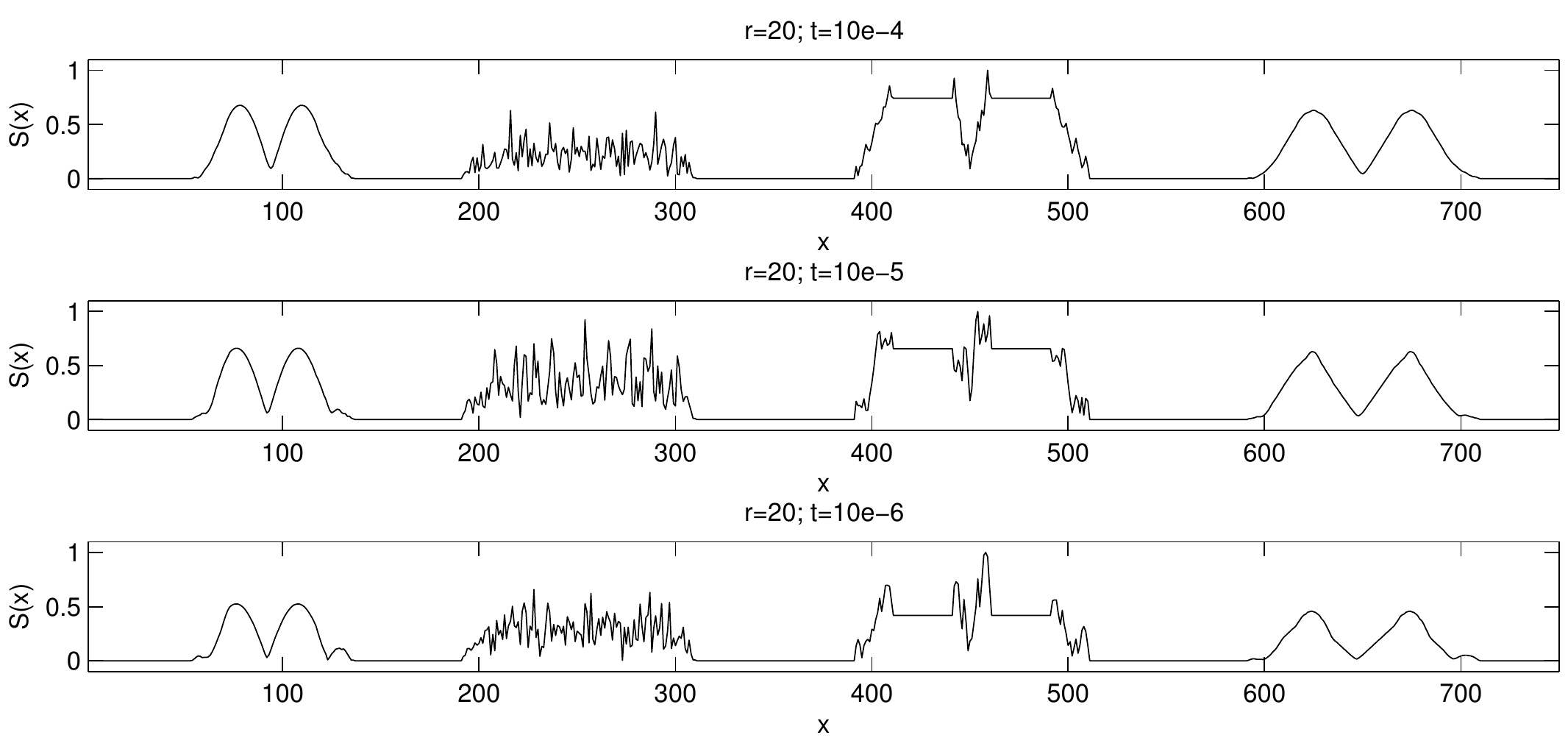}\\ 
        \hline
    \end{tabular}
    \begin{tabular}{|c|}
    	\hline
		\includegraphics[width=\textwidth]{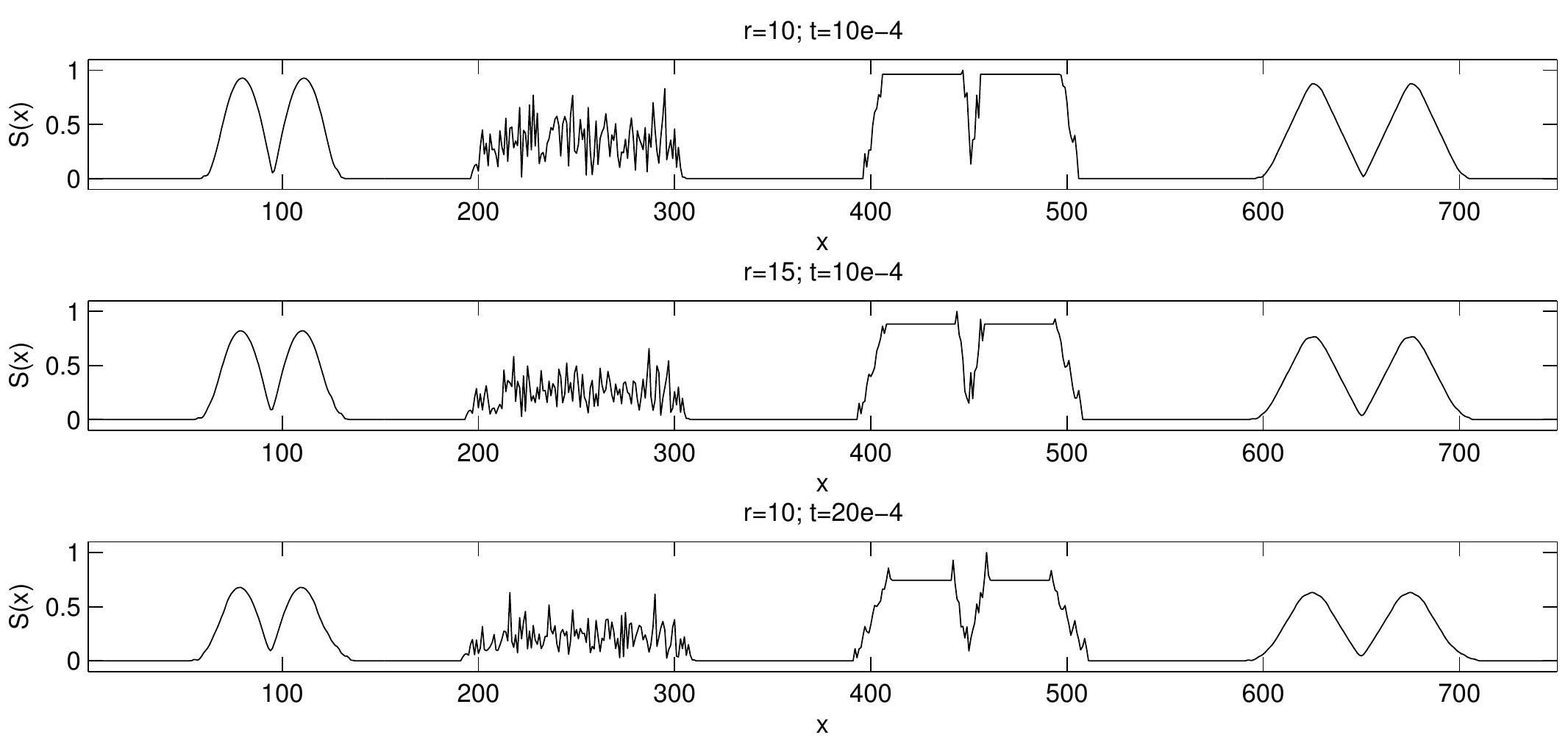}\\        
        \hline
	\end{tabular}
	\caption{Effect of Schroedinger kernels varying $t$ and $r$. On top, the original signal containing a smooth sinusoid, a random signal and a square wave. In the middle, $r$ is fixed in 20 and $t$ is defined as $10^{-4}$, $10^{-5}$ and $10^{-6}$. At the bottom, $t$ is fixed in $10^{-4}$ and $r$ is defined as 15, 20 and 25.}
	\label{fig:conv1D}
\end{figure}

Figures \ref{fig:conv1D} shows the operation of Schroedinger kernels over an artificial signal containing a sinusoidal, a square and a sawtooth wave as well as random signal. Those types of signal simulate practical situations where we have a combination of smooth and rough signal, as well as steps with smaller or higher magnitudes. 

The first thing noticed is the strong non-linear aspect of the kernel as expressed by the fluctuations arising in the transformation of a piece-wise smooth or even constant function. Second, the transform identified the magnitude of the jump in the signal intensity. Especially for high values of $r$ and small values of $t$ ($r=20,t=10^{-9}$ for instance), there was slightly more fluctuation when the jump in the square wave was larger. The transform also presented distinct behaviour when the jump was ascending or descending: in each sinusoidal wave or square ``tooth'', the fluctuation was more or less intensive depending on the direction of the jump and on the parameters $r$ and $t$ employed. Finally, as expected, smaller values of $t$ and larger values of $r$ caused more dramatic fluctuations in the transform and the variations in the original signal were more clearly captured. This fact confirms the multiscale aspect of the Schroedinger transform and will be fundamental to establish the method proposed here. In practice, all such characteristics of the transform emphasise the most prominent information in the image, which originally corresponds to variations in the texture pattern.

\begin{figure}[!htpb]
	\centering
	\begin{tabular}{cccc}
    \includegraphics[width=.22\textwidth]{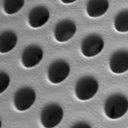} &
		\includegraphics[width=.22\textwidth]{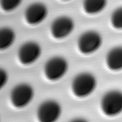} &
		\includegraphics[width=.22\textwidth]{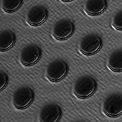} &
		\includegraphics[width=.22\textwidth]{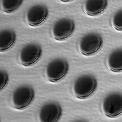}\\
		original & $t=1.1 \cdot 10^{-5}$ & $t=2.6 \cdot 10^{-5}$ & $t=3.6 \cdot 10^{-5}$\\
        \includegraphics[width=.22\textwidth]{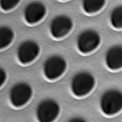} &
		\includegraphics[width=.22\textwidth]{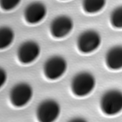} &
		\includegraphics[width=.22\textwidth]{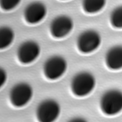} &
		\includegraphics[width=.22\textwidth]{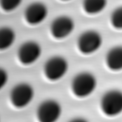}\\
		$t=5.6 \cdot 10^{-5}$ & $t=7.1 \cdot 10^{-5}$ & $t=9.1 \cdot 10^{-5}$ & $t=9.6 \cdot 10^{-5}$\\
        \hline
        \\
		\includegraphics[width=.22\textwidth]{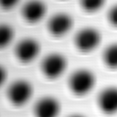} &
		\includegraphics[width=.22\textwidth]{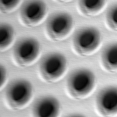} &
		\includegraphics[width=.22\textwidth]{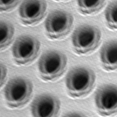} &
		\includegraphics[width=.22\textwidth]{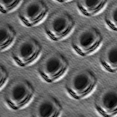}\\
		$t=1.1 \cdot 10^{-5}$ & $t=1.6 \cdot 10^{-5}$ & $t=3.1 \cdot 10^{-5}$ & $t=4.1 \cdot 10^{-5}$\\ 
        \includegraphics[width=.22\textwidth]{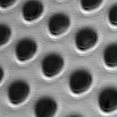} &
		\includegraphics[width=.22\textwidth]{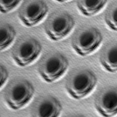} &
		\includegraphics[width=.22\textwidth]{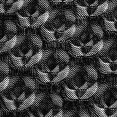} &
		\includegraphics[width=.22\textwidth]{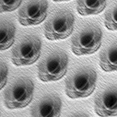}\\
		$t=5.6 \cdot 10^{-5}$ & $t=7.1 \cdot 10^{-5}$ & $t=9.1 \cdot 10^{-5}$ & $t=9.6 \cdot 10^{-5}$\\ 
	\end{tabular}
	\caption{Examples of images operated with different Schroedinger kernels. In the top left the original image is showed. The first group shows 7 images operated with kernel size $r=5$ and the second group shows 8 images operated with $r=10$. For each image the $t$ parameter used is showed below.}
	\label{fig:filterSchroendinger}
\end{figure}

Similar effect is observed on a random signal (Figure \ref{fig:conv1D}) and on a grey-level image in Figure \ref{fig:filterSchroendinger}. In the last case, similar to what happened with 1D signals, it can be observed the impact of the kernel size $r$ and parameter $t$. Whereas smaller values of $r$ or larger values of $t$ tend to smooth the image without a dramatic change in the pattern structure, the opposite situation causes the arising of notably non-linear patterns that substantially change the relation between pixels within a neighbourhood in the original image, expressing variations in the image both in terms of magnitude and extension.
\color{black}

The image descriptors $D$ here proposed are obtained at the end by computing central moments of the distribution of values in the Schroedinger transform. First, it was empirically observed that the best range for $t$ lies between $10^{-6}$ and $10^{-4}$ and the moments are computed for each value of $t$ within this range and concatenated. Therefore, they are represented by a compact notation:
\[
	D_t^r = \bigcup_{i=1}^{100}\bigcup_{m=1}^{M}\mu_{10^{-4}i,r}^{m},
\]
where $\mu_{t,r}^m$ is the central moment:
\[
	\mu_{t,r}^m = \sum_{x}{(x-\overline{h_{t,r}(x)})^m h_{t,r}(x) dx},
\]
being $h_{t,r}(x)$ the histogram of the Schroedinger transform $S$ with kernel size $r$ and the parameter $t$. Besides the underlying invariance of this type of moments, here they capture in a simple yet representative way all the most important statistical properties of the transform, ensuring in this way the computation of a complete set of features capable of efficiently quantifying the texture image.

\section{Experiments}

The proposed method is assessed over three databases of textures used as benchmark, namely, Outex \cite{OjalaMPVKH02}, UIUC \cite{KA11} and USPTex \cite{BackesB10}. Outex is a set of 68 texture images and here each one is divided into 20 non-overlapping images with 128$\times$128 pixels, resulting in 1360 images divided into 68 classes. In a similar way, UIUC is composed by 40 classes with 25 images in each one and USPTex contains 191 classes, each one with 12 images. USPTex and Outex images are coloured and to be used here they are previously converted to grey-levels.

The proposed Schroedinger descriptors are applied to the classification of the databases and the ratio of images correctly classified is compared to other well-known texture descriptors in the literature, to know, Grey-Level Co-occurrence Matrix (GLCM) \cite{H67}, Fourier \cite{GW02}, multifractals \cite{XJF09}, Local Binary Patterns (LBP) \cite{OPM02}, LBP+VAR \cite{OPM02}, and MR8 \cite{VZ05}.

These descriptors are submitted to a Principal Component Analysis (PCA) \cite{DH00} to reduce dimensionality and discard less significant features. Finally the most significant PCA scores are applied to a Linear Discriminant Analysis \cite{DH00} on a 10-fold cross validation scheme to provide the classes of each sample.

For a more complete evaluation of the potential of the proposed method in real-world problems, it was applied to the identification of plant species from the Brazilian \textit{flora} using images from their respective leaves collected by a commercial scanner. The database, named 1200Tex and described in \cite{CMB09} contains 20 classes (species) and 60 windows with resolution 128$\times$128 in each class. 

\section{Results and Discussion}

The following figures and tables exhibit the results for the application of the proposed descriptors to the classification of the benchmark databases, as well as when compared to other well known texture descriptors in the literature.

The first test was carried out to observe the performance of the descriptors when the size $r$ of the kernel and the number $M$ of moments are varied. Table \ref{tab:CR_best} shows the success rates (percentage of images correctly classified) for each database, and using each combination of $M$ and $r$. The success rate fluctuates in a different way for each database. The lack of any pattern is explained by the heterogeneity of each data set, which causes different values for the parameters to be more adequate for a specific sub-part of the database. A clearer conclusion drawn from the tables is that values of $r$ between 2 and 8 present the best performance, while the ideal number of moments is largely dependent on the database. In practice, a training set can be separated to determine the ideal value for this parameter. 
\begin{table}[!htpb]
	\centering
	\caption{Best success rates for the optimum values of $at$ for the databases (a) UIUC, (b) Outex, (c) USPTex and (d) 1200Tex}	
	\footnotesize
	\begin{tabular}{c}
		\begin{tabular}{c c}
			\begin{tabular}{|c| c c c c|}
				\hline
				r			& M=5  	& M=10 	& M=15 	& M=20\\
				\hline
				2			& 82.40 & 84.50 & 83.40 & 83.10\\
				4			& 88.70 & 87.80 & 86.70 & 85.60\\
				6 			& 89.40 & 87.50 & 86.20 & 85.80\\
				8			& 89.30 & 87.20 & 85.60 & 83.90\\
				10        	& 88.40 & 85.90 & 84.60 & 84.10\\
				12          & 87.60 & 84.50 & 83.50 & 82.50\\
				14       	& 87.00 & 83.90 & 81.70 & 80.80\\
				16       	& 86.90 & 83.70 & 81.20 & 80.70\\
				18       	& 85.10 & 82.40 & 79.70 & 78.50\\
				20       	& 83.30 & 80.50 & 78.50 & 76.90\\
				\hline
			\end{tabular} &
			\begin{tabular}{|c| c c c c|}
				\hline
				r			& M=5 	& M=10 	& M=15 	& M=20\\
				\hline
				2			& 78.16 & 79.12 & 79.56 & 80.07\\
				4			& 83.23 & 83.60 & 82.94 & 82.79\\
				6 			& 84.12 & 83.90 & 83.97 & 83.90\\
				8			& 82.72 & 82.21 & 82.06 & 81.69\\
				10        	& 83.60 & 83.90 & 83.31 & 83.23\\
				12          & 81.62 & 81.32 & 81.47 & 81.47\\
				14       	& 82.43 & 82.35 & 81.76 & 81.69\\
				16       	& 81.69 & 81.47 & 81.25 & 81.54\\
				18       	& 82.13 & 82.35 & 82.28 & 82.35\\
				20       	& 79.78 & 79.93 & 79.63 & 79.63\\			
				\hline
			\end{tabular}\\
			(a) UIUC & (b) Outex\\
		\end{tabular}\\~\\
		\begin{tabular}{c c}				
			\begin{tabular}{|c| c c c c|}
				\hline
				r			& M=5 	& M=10 	& M=15 	& M=20\\
				\hline
				2			& 77.36 & 80.24 & 80.80 & 80.50\\
				4			& 86.82 & 88.30 & 87.65 & 87.26\\
				6 			& 88.09 & 88.53 & 87.30 & 87.04\\
				8			& 87.48 & 86.78 & 85.95 & 85.78\\
				10        	& 86.65 & 85.60 & 84.38 & 83.55\\
				12          & 85.82 & 83.94 & 83.16 & 82.68\\
				14       	& 85.04 & 83.42 & 82.37 & 81.54\\
				16       	& 83.20 & 81.15 & 80.41 & 79.45\\
				18       	& 82.50 & 80.63 & 80.63 & 80.63\\
				20       	& 80.45 & 78.01 & 77.05 & 76.35\\			
				\hline
			\end{tabular}  &
			
	\begin{tabular}{|c| c c c c|}
		\hline
		r			& M=5 	& M=10 	& M=15 	& M=20\\
		\hline
		2			& 71.75 & 72.58 & 72.83 & 71.83\\
		4			& 81.00 & 80.25 & 79.58 & 79.33\\
		6			& 82.58 & 80.41 & 79.67 & 79.25\\
		8			& 80.83 & 79.08 & 78.17 & 77.92\\
		10			& 81.42 & 80.25 & 79.42 & 78.92\\
		12			& 80.83 & 79.58 & 78.83 & 78.67\\
		14			& 80.25 & 78.33 & 77.67 & 76.92\\
		16			& 78.25 & 77.00 & 76.17 & 76.08\\
		18			& 77.92 & 76.75 & 75.83 & 75.33\\
		20			& 75.58 & 74.67 & 73.00 & 73.25\\
		\hline
	\end{tabular}\\
			(c) USPTex & (d) 1200Tex\\
		
	\end{tabular}			
	\end{tabular}		
	\label{tab:CR_best}
\end{table}

Table \ref{tab:CR_best_comp} shows a comparison between the success rates (and respective errors) of the other methods in the literature, comparing their performance to the proposed approach. The Schroedinger descriptors outperformed even methods well known for the good performance methods like LBP+VAR and textons (MR8). Another interesting point is that the good performance of the proposed method is made even more evident on UIUC and USPTex. Such advantage in the classification of images like those from USPTex is a relevant achievement considering the challenge of that data set, which besides being highly heterogeneous, comprises a larger number of samples and classes than the other databases.
\begin{table}[!htpb]
	\centering
	\caption{Best success rates for the optimum values of $r$ and $M$.}	
	\begin{tabular}{c c c c}
		\hline
      	Method	 		& UIUC				& Outex 			& USPTex\\
        \hline
		\hline
		Fourier			& 64.00$\pm$0.03	& 82.21$\pm$0.02   	& 71.16$\pm$0.03\\
		GLCM			& 58.70$\pm$0.03	& 68.23$\pm$0.05   	& 74.22$\pm$0.02\\
		Multifractals 	& 82.40$\pm$0.03	& 75.07$\pm$0.03   	& 68.76$\pm$0.03\\
		LBP				& 77.40$\pm$0.05	& 74.63$\pm$0.05	& 79.98$\pm$0.03\\		
		LBP+VAR        	& 83.90$\pm$0.02    & 76.03$\pm$0.04   	& 84.52$\pm$0.04\\
		MR8            	& 84.70$\pm$0.02    & 71.54$\pm$0.04   	& 42.37$\pm$0.05\\
		\hline
		Proposed		& 89.40$\pm$0.03   	& 84.12$\pm$0.03   	& 88.53$\pm$0.01\\
		\hline
	\end{tabular}
	\label{tab:CR_best_comp}
\end{table}

\begin{figure}[!htpb]
	\centering
		\begin{tabular}{c}
			Outex \\
		\end{tabular} \\	
		\begin{tabular}{cc}
			\includegraphics[width=.5\textwidth]{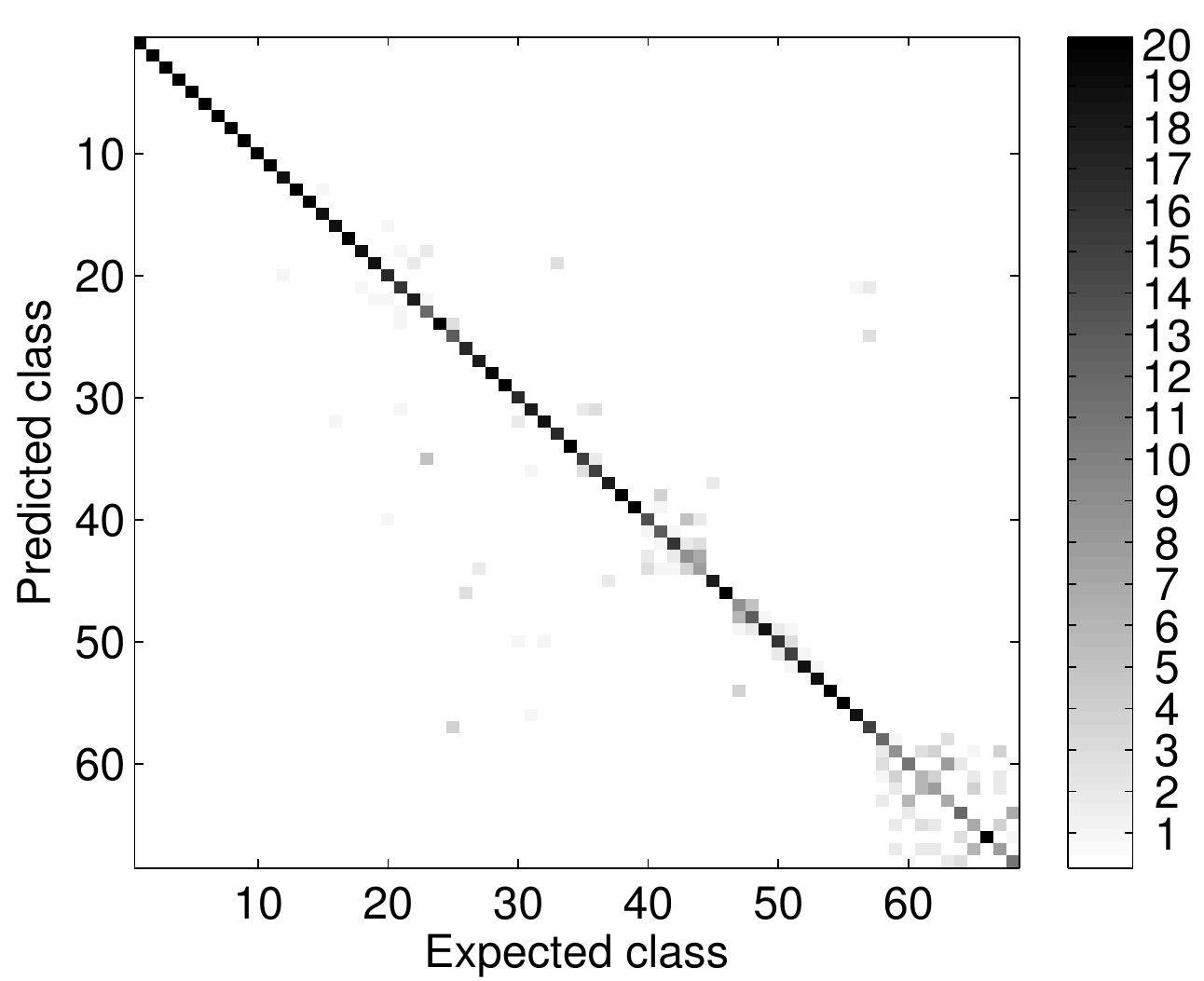} &
	      	\includegraphics[width=.5\textwidth]{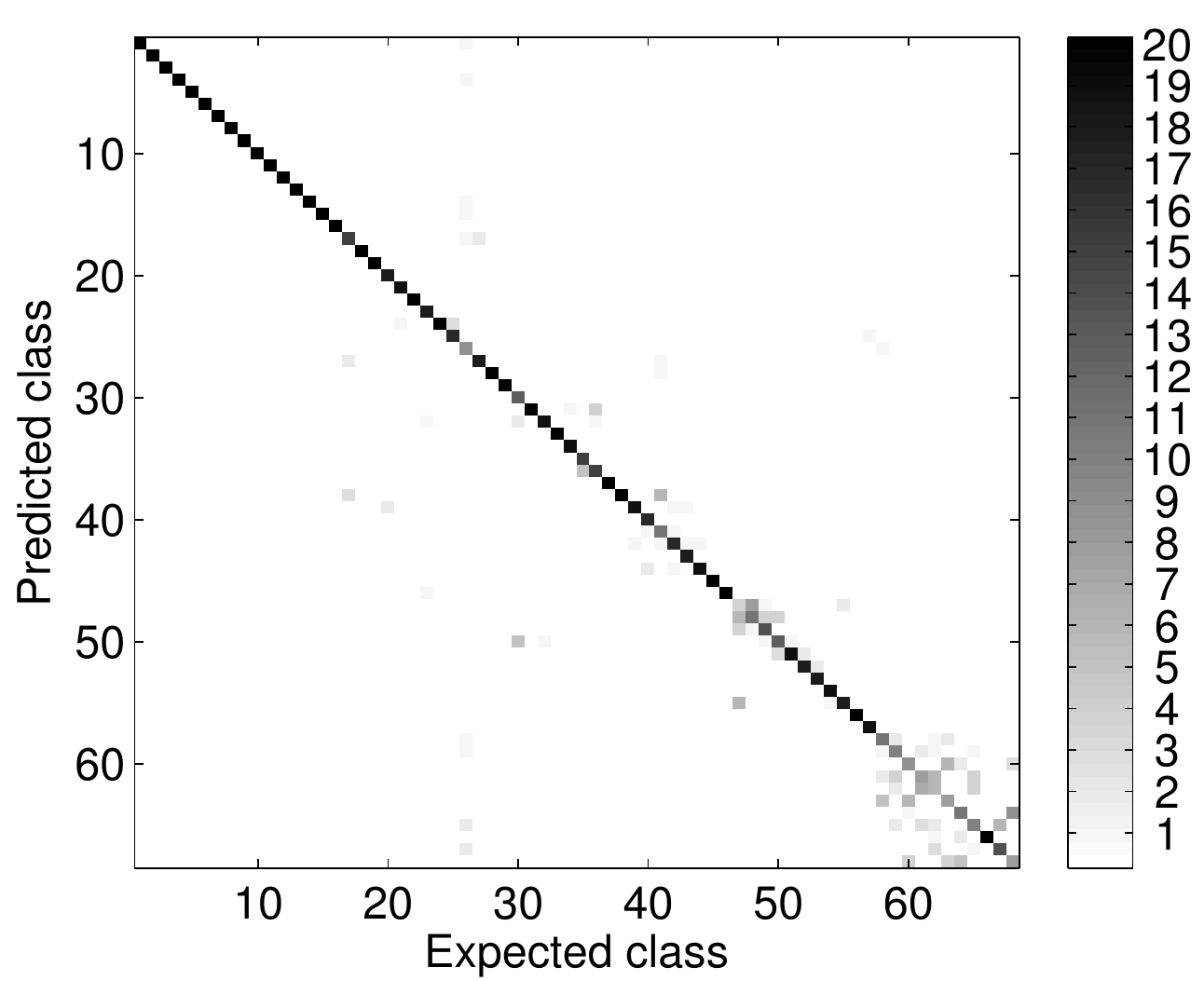}\\
			VZ-Joint. & Schroedinger. \\
		\end{tabular} \\
		\begin{tabular}{c}
			UIUC \\
		\end{tabular} \\
		\begin{tabular}{cc}
			\includegraphics[width=.5\textwidth]{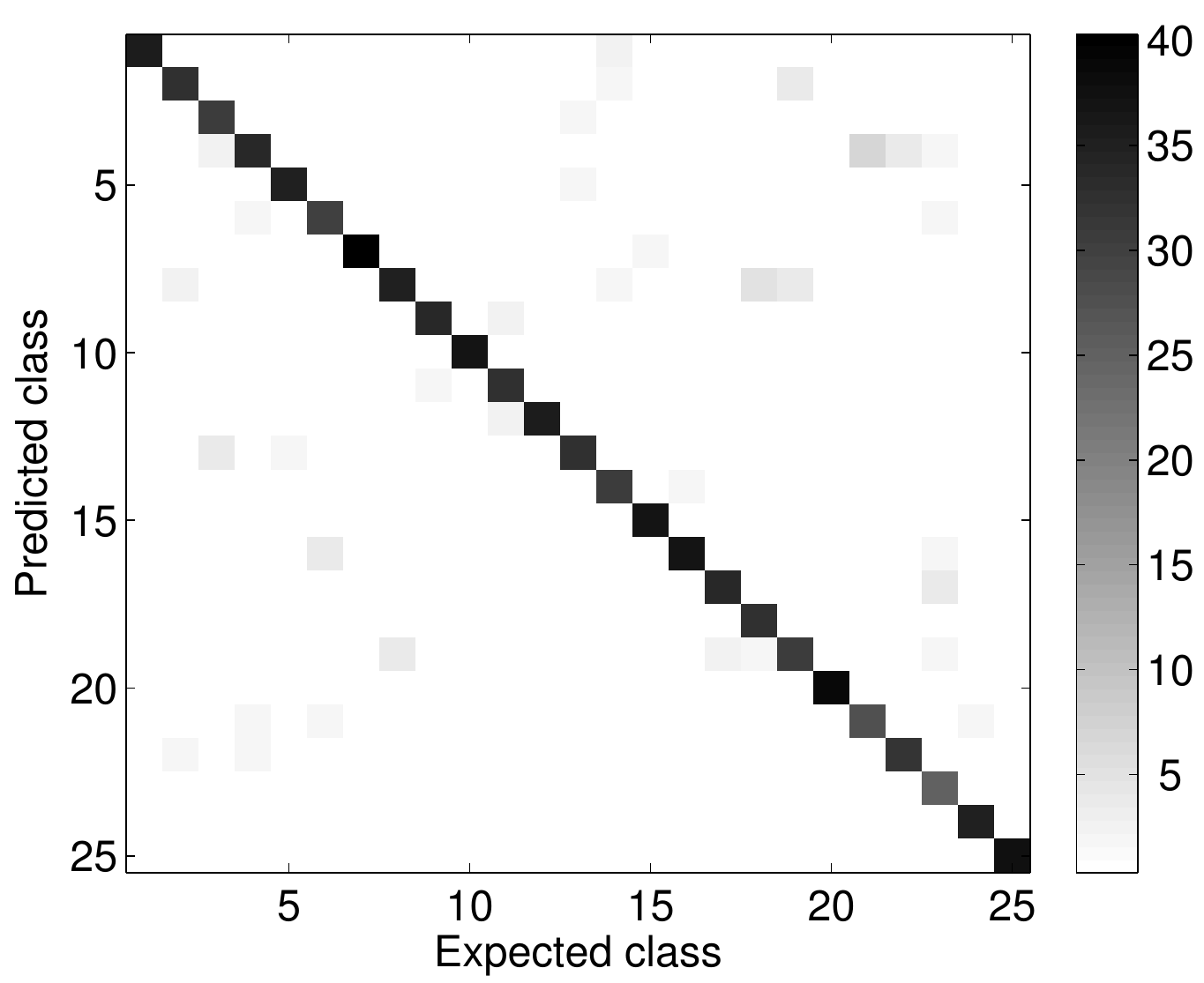} &
	      	\includegraphics[width=.5\textwidth]{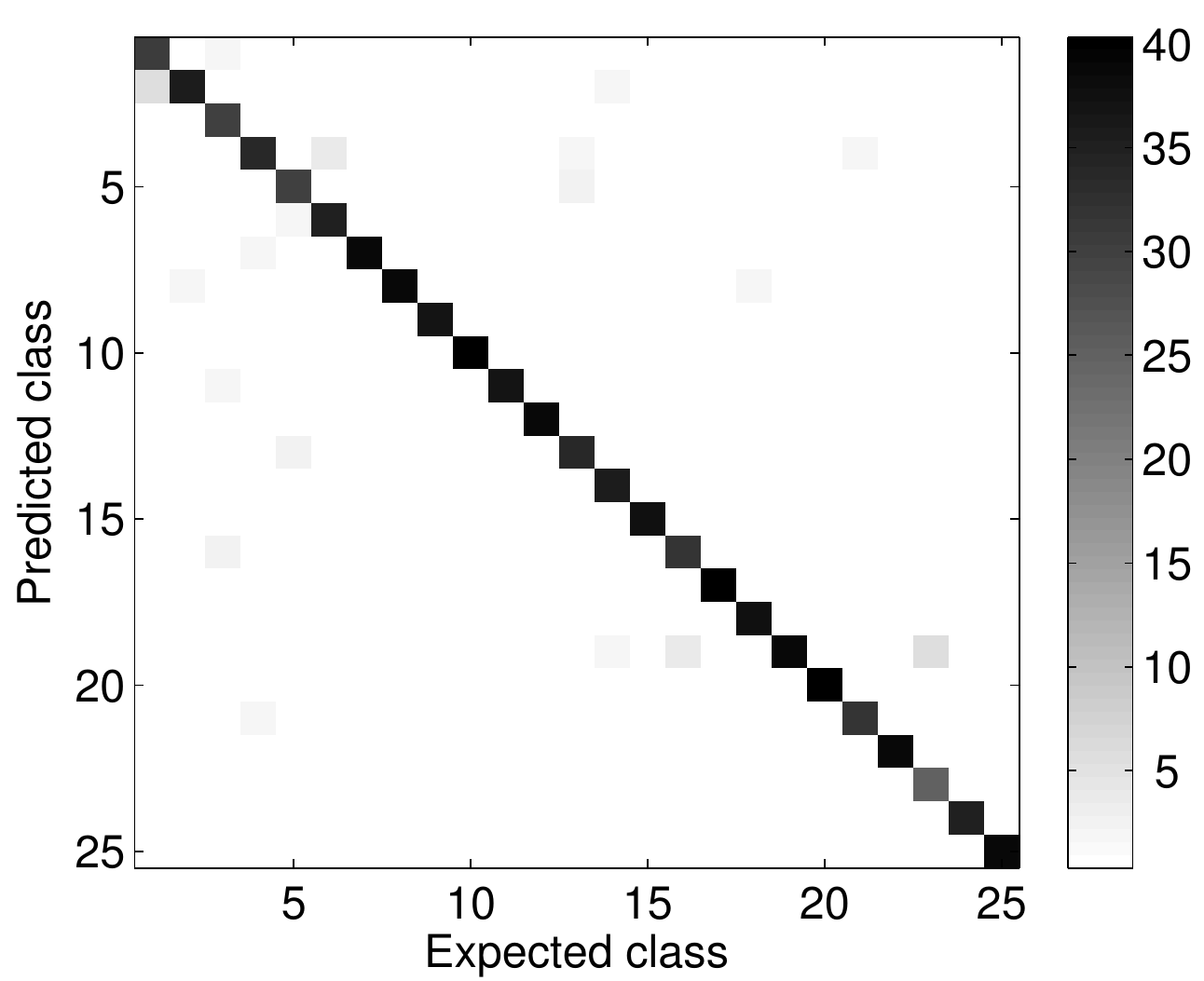}\\
			MR8. & Schroedinger. \\
		\end{tabular} \\		
		\begin{tabular}{c}
			USPTex \\
		\end{tabular} \\
		\begin{tabular}{cc}
			\includegraphics[width=.5\textwidth]{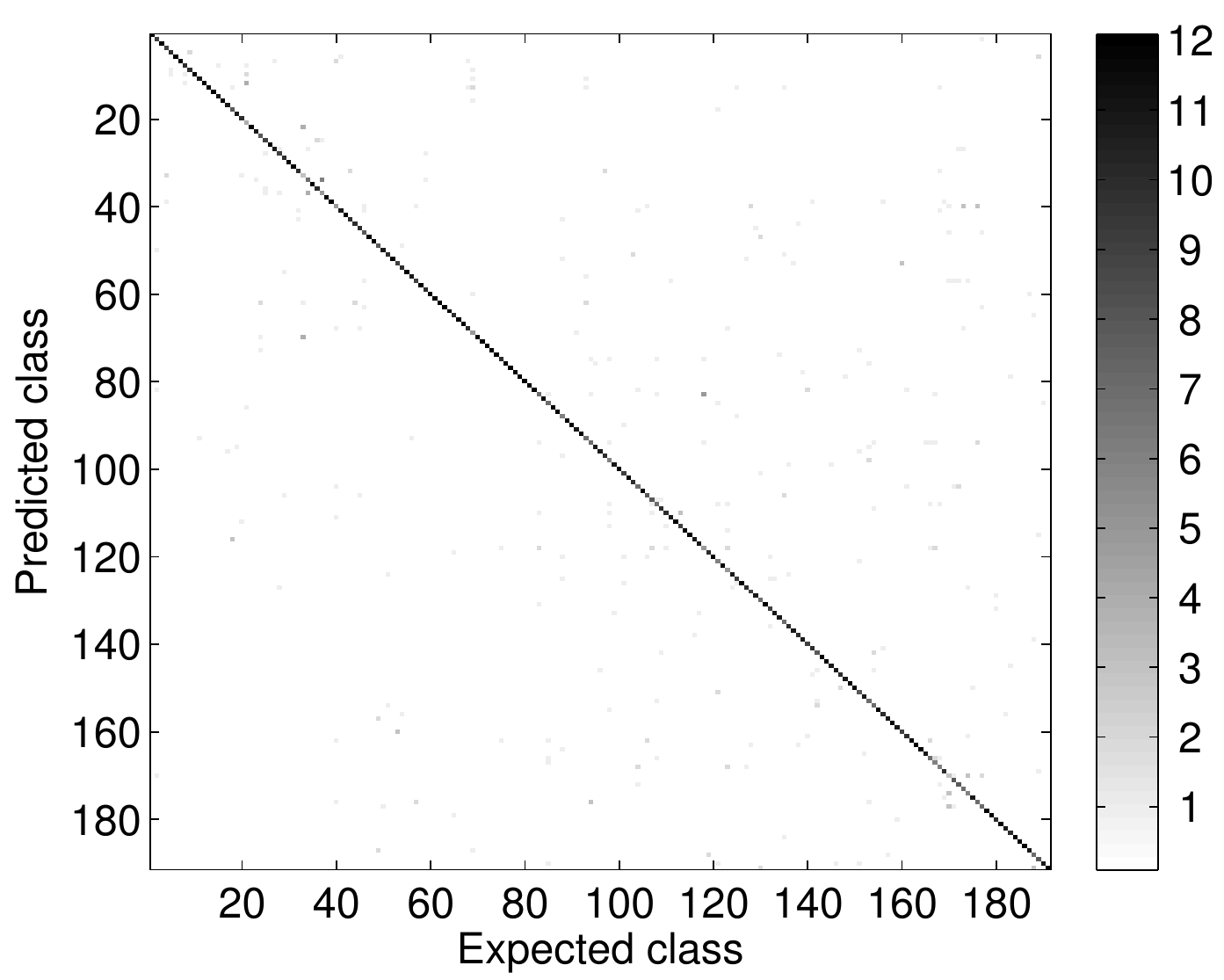} &
	      	\includegraphics[width=.5\textwidth]{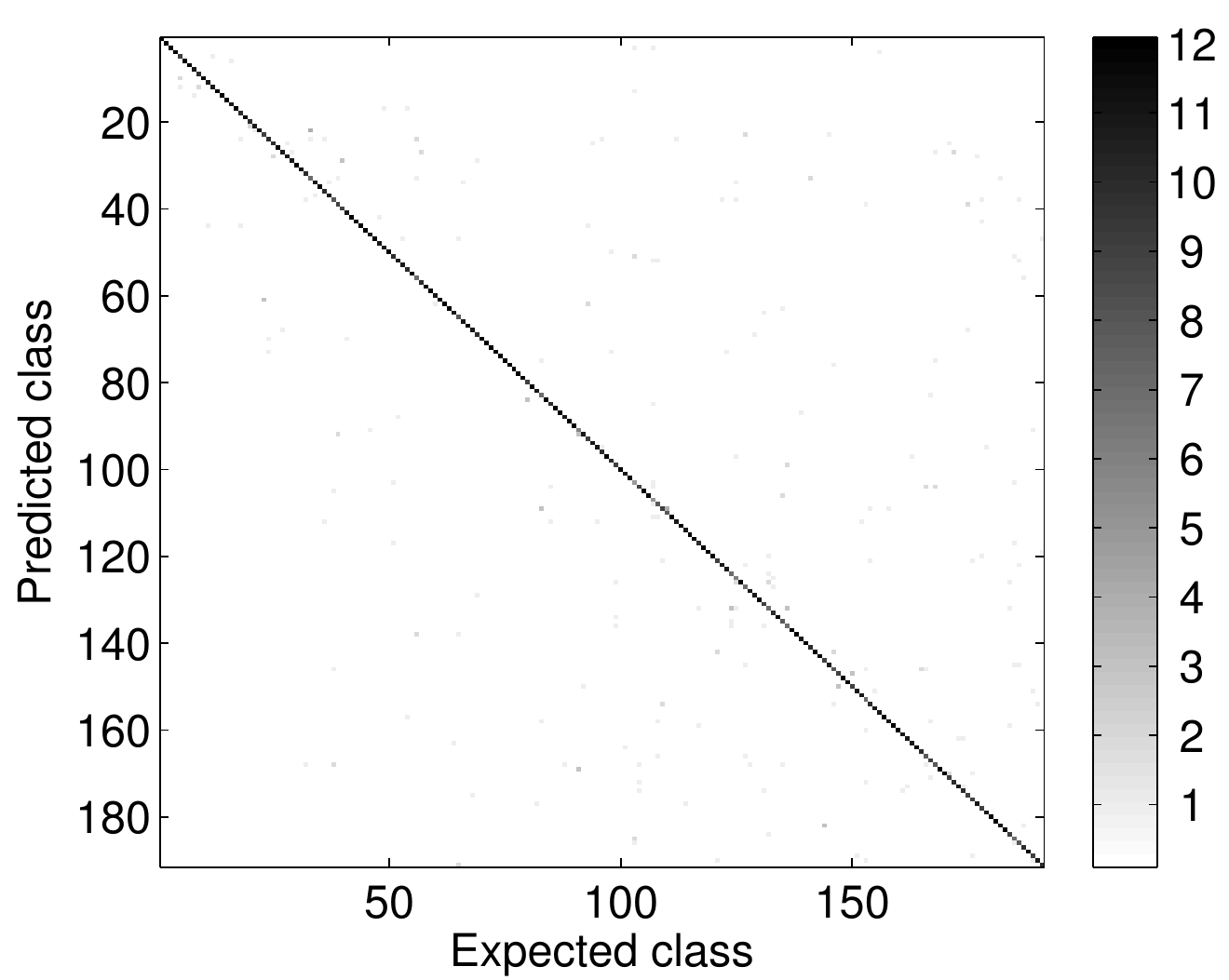}\\
			LBP+VAR. & Schroedinger. \\
		\end{tabular} \\				
	\caption{Confusion matrices of Schroedinger descriptors compared to the second best approaches in Table \ref{tab:CR_best_comp}.}
    \label{fig:CM}                                  
\end{figure}

Figure \ref{fig:CM} expresses the class-wise performance of the compared methods by showing the confusion matrices of Schroedinger compared to their counterpart when the images are classified by other methods (those presenting the second best performance in Table \ref{tab:CR_best_comp}). The easiest way of verifying which one is the best method with regard to the class-wise behaviour is by counting the number of grey points outside the diagonal, as well as the number of light points on the diagonal. Such representation is also helpful to find out the most recommended method to discriminate between two or more particular classes.

\begin{figure}[!htpb]
	\centering	
	\begin{tabular}{cc}
		\includegraphics[width=.5\textwidth]{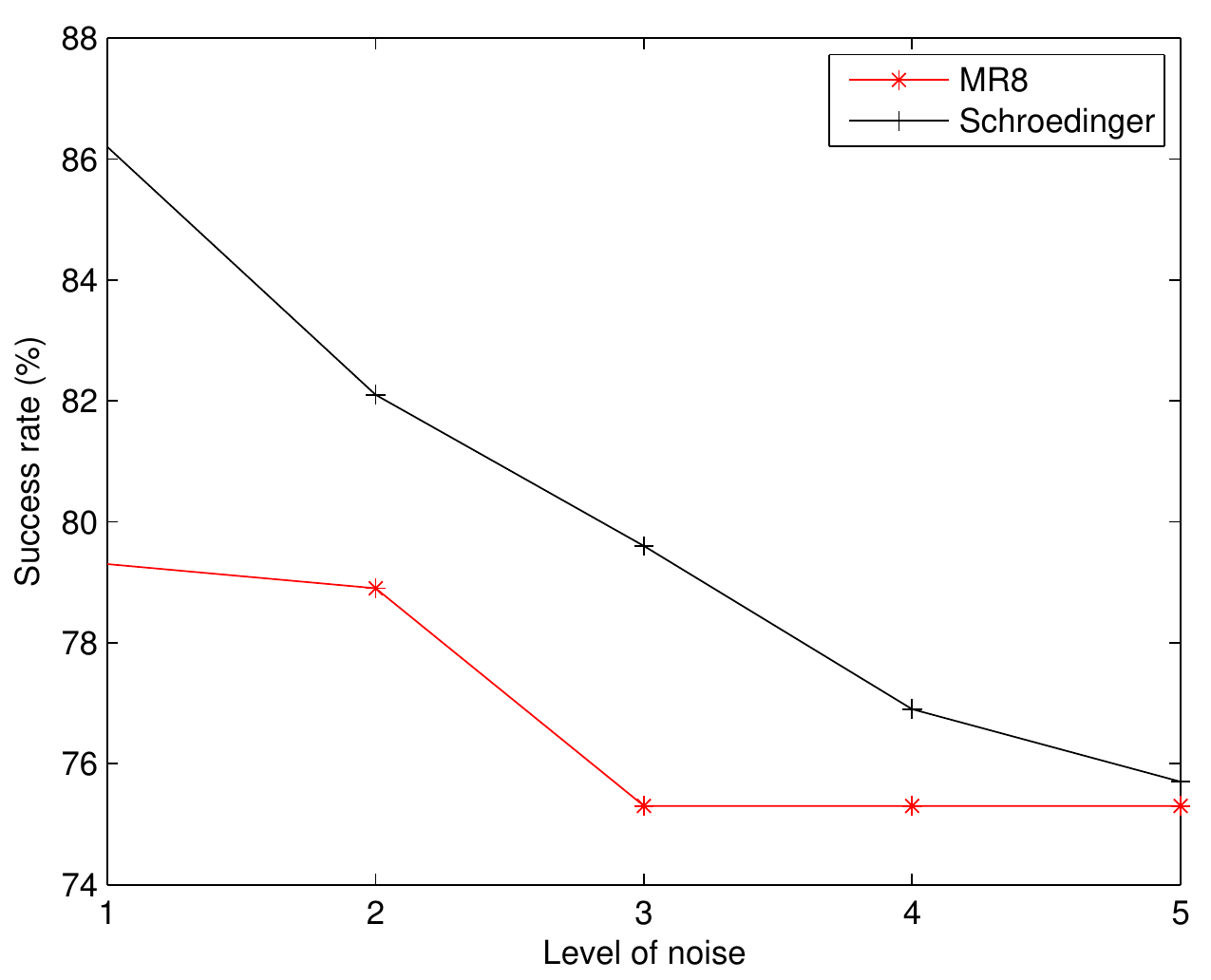} &
		\includegraphics[width=.5\textwidth]{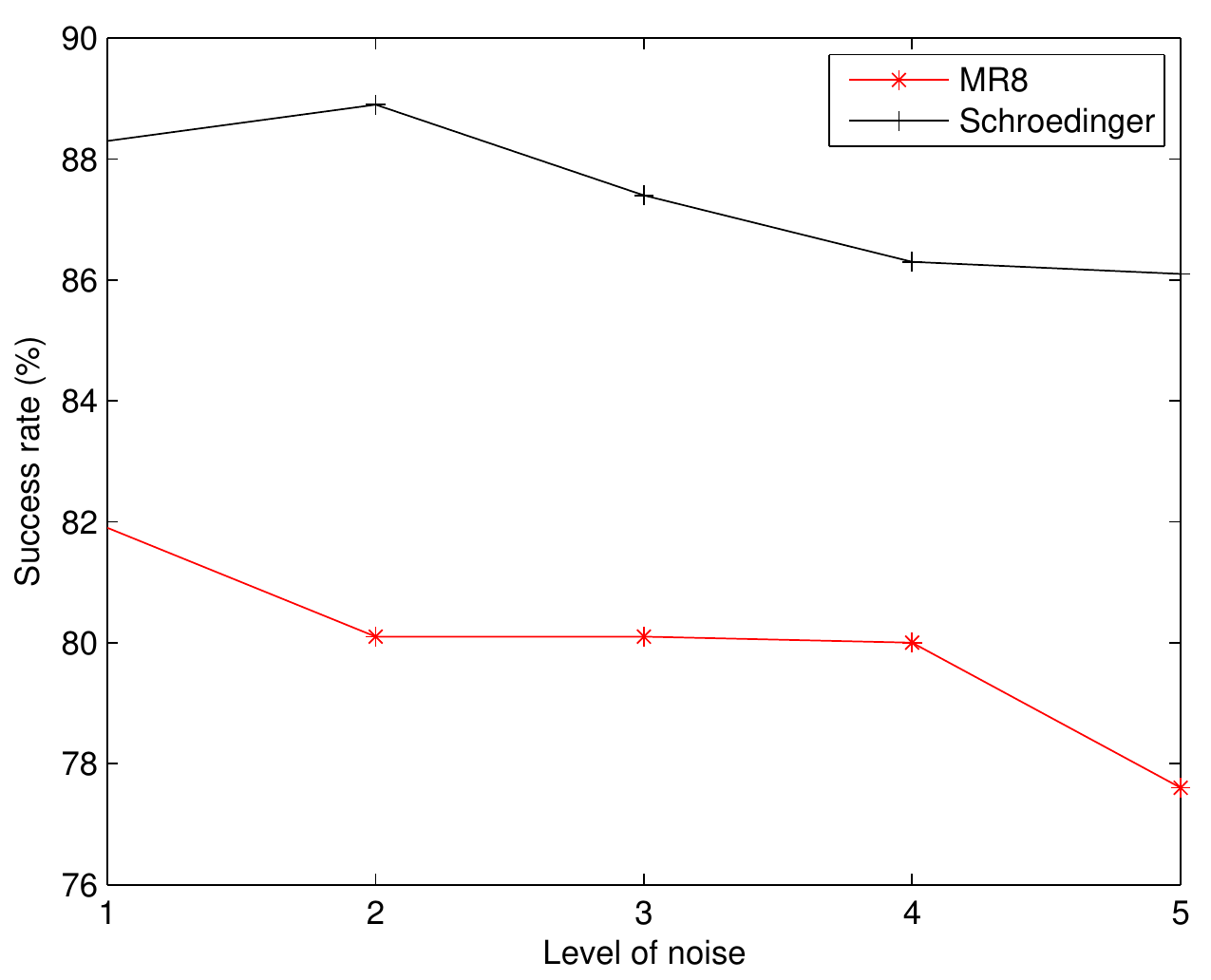}\\
		Gaussian. & Salt \& Pepper. \\
	\end{tabular} \\		
	\caption{Success rates in the presence of Gaussian and ``Salt \& Pepper'' noise.}
	\label{fig:noise}                                  
\end{figure}

Figure \ref{fig:noise} shows the performance of the proposed features when the images in UIUC database are subjected to two types of random noises (Gaussian and ``salt \& pepper''). The graphs exhibit clear advantage of the Schroedinger descriptors, mainly in the presence of ``salt \& pepper'' noise.

The last experiment was carried out using the 1200tex leaf database. Table \ref{tab:CR_best} (d) shows the success rates (percentage of images correctly classified) using each combination of $M$ and $r$ and Table \ref{tab:CR_plants} exhibits the success rates achieved by the Schroedinger descriptors compared with the other approaches employed in this task. As it occurred for the benchmark data sets, the proposed method outperformed all the other presented descriptors. Our method was significantly better than other texture descriptors such as LBP+VAR and MR8.

\begin{table}[!htpb]
	\centering
	\caption{Comparison of the proposed method and others in the 1200Tex database. For the proposed method, the best success rates for the optimum values of $r$ and $M$ were used.}	
	\begin{tabular}{c c}
		\hline
      	Method	 		& Success rate (\%)\\
        \hline
		\hline
		Fourier			& 70.25$\pm$0.04\\
		GLCM			& 52.67$\pm$0.04\\
		Multifractals 	& 57.92$\pm$0.04\\
		LBP				& 68.83$\pm$0.05\\		
		LBP+VAR        	& 73.25$\pm$0.03\\
		MR8            	& 53.42$\pm$0.04\\
		\hline
		Proposed		& 82.58$\pm$0.03\\
		\hline
	\end{tabular}
	\label{tab:CR_plants}
\end{table}

\begin{figure}[!htpb]
	\centering
		\begin{tabular}{cc}
			\includegraphics[width=.5\textwidth]            {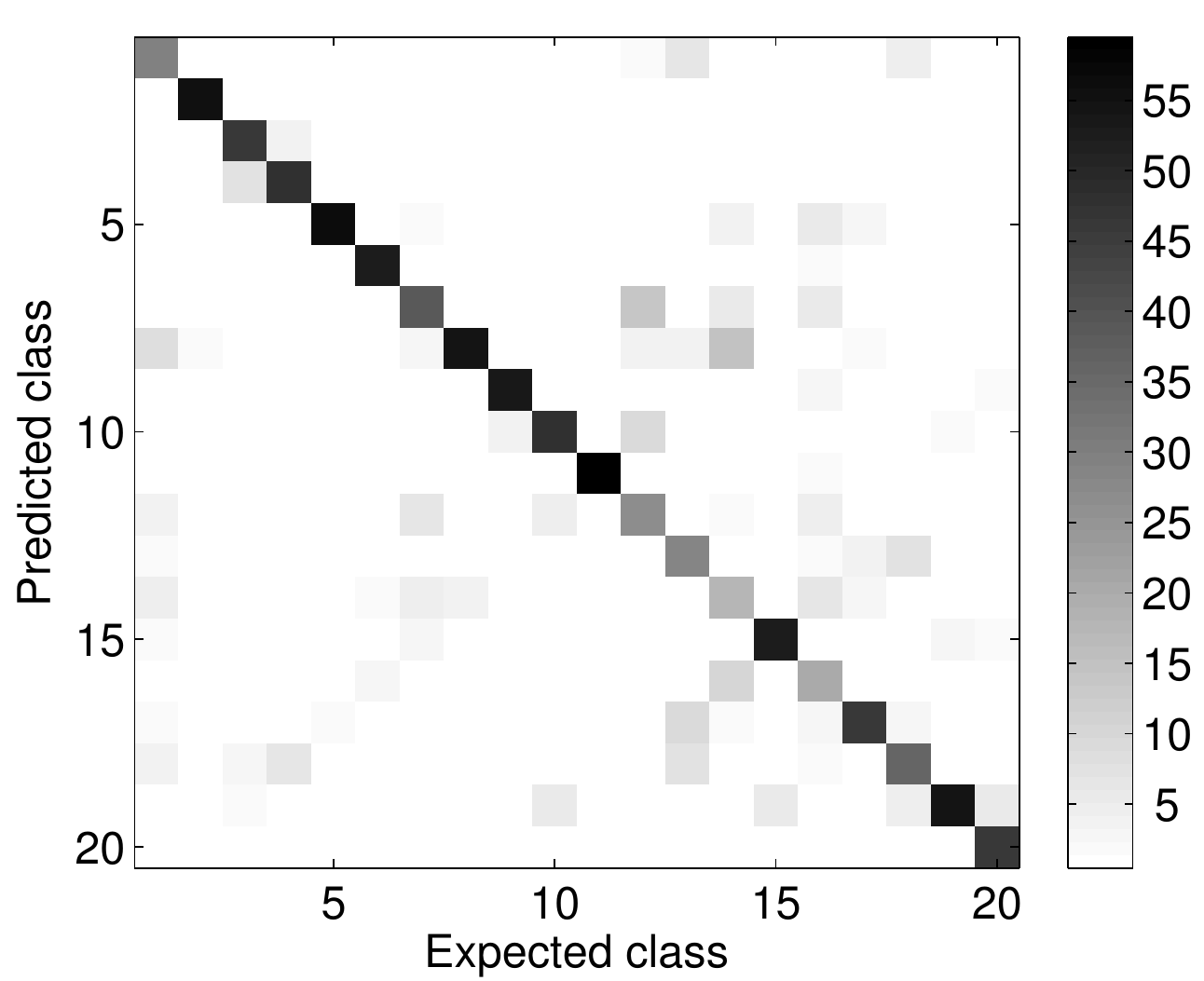} &
	      	\includegraphics[width=.5\textwidth]{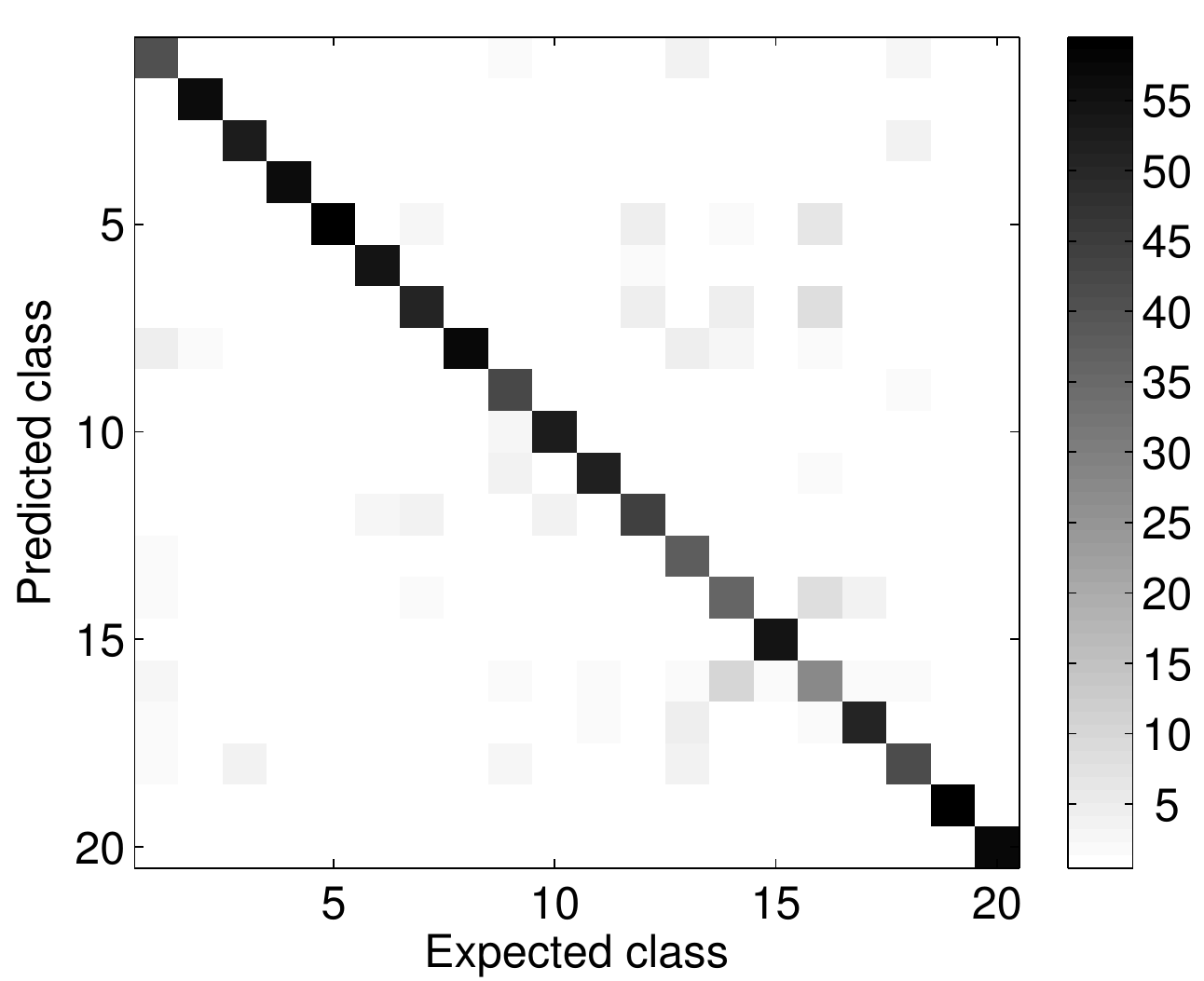}\\
			LBP+VAR. & Schroedinger. \\
		\end{tabular} \\				
	\caption{Confusion matrices of Schroedinger descriptors compared to the second best approach in Table \ref{tab:CR_plants}.}
    \label{fig:CM_plants}                                  
\end{figure}

\begin{figure}[!htpb]
	\centering
		\begin{tabular}{ccc}
			\includegraphics[width=.2\textwidth]{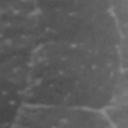} &
	    \includegraphics[width=.2\textwidth]{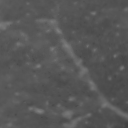} &
			\includegraphics[width=.2\textwidth]{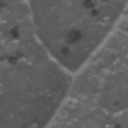} \\
			\includegraphics[width=.2\textwidth]{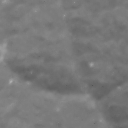} &
	    \includegraphics[width=.2\textwidth]{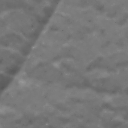} &
			\includegraphics[width=.2\textwidth]{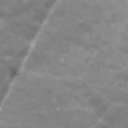} \\
			\includegraphics[width=.2\textwidth]{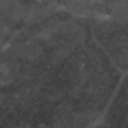} &
	    \includegraphics[width=.2\textwidth]{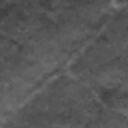} &
			\includegraphics[width=.2\textwidth]{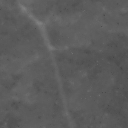} \\						
		\end{tabular} \\				
		\caption{Samples from 1200Tex database illustrating discontinuities in the texture image.}
    \label{fig:examples_plants}                                  
\end{figure}

To better understand the mechanisms involved in the proposed descriptors and identify in practice what kind of texture image is more appropriate for an analysis using the method, especially in comparison to other texture descriptors, Figure \ref{fig:CM_plants} shows the respective confusion matrices. Three species (7, 12, and 14) should be more carefully analysed as they present the most significant discrepance in favour of Schroedinger method (85\%$\times$65\%, 75\%$\times$45\%, and 60\%$\times$30\%, respecively). By observing samples in these classes it can be noticed that those textures also are characterised by the discontinuities pointed out in the presentation of the proposed method. Figure \ref{fig:examples_plants} exhibits a few samples exemplifying such behaviour.

\subsection{Discussion}

Generally speaking, the proposed method provided descriptors capable of classifying the benchmark databases with higher precision among all the seven compared methods, even when the databases present challenging properties, like the high number of classes or a high/low dissimilarity intra/inter-class, or when the images are subjected to the influence of random noise. Such advantage can be explained by understanding the way that Schroedinger transform works over the texture. It essentially operates at different scales (by setting appropriate values of $r$) and highlights discontinuities on the image. Nevertheless, unlike classical techniques such as high-pass filters for example, the global information enclosed by the kernel $K(b,a)$, which takes into account all the possible paths, allows to detect not only the place where such discontinuities emerge in the image, but also the magnitude of the jump in the pixel intensity, the size of homogeneous regions limited by such edges and even the direction followed by the edge. When the moments are extracted, the highlighted edges make the homogeneous regions more evident whereas the multiscale operation makes the descriptors less susceptible to noise and small neighbourhood variations, giving rise to more reliable features.

Considering the above, the method presented here is especially recommended for the analysis of textures that, besides the usual presence of recognizable statistical patterns, also can be described in terms of their discontinuities. Despite the fact that this type of image can be easily found, mainly in the analysis of heterogeneous materials, the literature on pattern recognition in grey-level images uses to model these features as if they were two independent characteristics: textures and shapes (or edges), respectively. The analysis proposed by the Schroedinger descriptors works in practice like a joint distribution of texture and heterogeneity descriptors, allowing in this way a more complete and robust representation of the image. Such robustness was confirmed here by the interesting results achieved over the benchmark databases, including situations with the presence of noise.

Regards to the leaf plant database (1200Tex), it is possible to say that the proposed method confirmed the positive expectations in this practical situation and showed its value as an automatic solution to help taxonomists in such a complex task as the identification of plant species. This test also illustrates a typical example where the positioning of discontinuities along the texture is a potential candidate to provide meaningful information about the object being analysed. Particularly, the surface of leaves are commonly permeated by veins at different levels of scale (primary and secondary), whose distribution is known to have important role in distinguishing plant species \cite{PPFVOB05,BPFC08}.

\section{Conclusions}

This work proposed to apply the discrete Schroedinger transform to provide descriptors of grey-level texture images. The obtained features were employed to classify three classical databases of textures and outperformed other texture descriptors in the literature, such as LBP+VAR, MR8 and others. A similar test was accomplished to assess the influence of random noise over the images and again the proposed features presented the best performance. Such good result was expected from the individual operation of Schroedinger transform. While this transform can act as a conventional multiscale operator, at the same time it also highlights particular edges making the images less sensitive to noises and artifacts, and quantifying the degree of homogeneity within a range of scales in the transformed image.

More than the numerical advantage, the high percentage of images correctly classified in so large and heterogeneous databases suggests more attention to the Schroedinger transform as an auxiliary operation for the extraction of image descriptors and encourages its application in real-world problems where a precise description of texture images plays important role.

\section*{Acknowledgements}
J. B. F. gratefully acknowledges the financial support of FAPESP Proc. 2012/19143-3. O.M.B. acknowledges support from CNPq (Grants \#307797/2014-7 and \#484312/2013-8) and FAPESP (Grant \#14/08026-1).


\end{document}